# Agile, Autonomous Spacecraft Constellations with Disruption Tolerant Networking to Monitor Precipitation and Urban Floods

Sreeja Roy-Singh[1], Alan P. Li[1,*], Vinay Ravindra[1], Roderick Lammers[2,*], Marc Sanchez Net[3]
[1]NASA Ames Research Center, BAER Institute, Moffet Field, CA
[2]University of Georgia, Athens, GA
[3]NASA Jet Propulsion Laboratory, California Institute of Technology, Pasadena, CA

*Abstract*— **Fully re-orientable small spacecraft are now supported by commercial technologies, allowing them to point their instruments in any direction and capture images, with short notice. When combined with improved onboard processing, and implemented on a constellation of inter-communicable satellites, this intelligent agility can significantly increase responsiveness to transient or evolving phenomena. We demonstrate a ground-based and onboard algorithmic framework that combines orbital mechanics, attitude control, inter-satellite communication, intelligent prediction and planning to schedule the time-varying, re-orientation of agile, small satellites in a constellation. Planner intelligence is improved by updating the predictive value of future space-time observations based on shared observations of evolving episodic precipitation and urban flood forecasts. Reliable inter-satellite communication within a fast, dynamic constellation topology is modeled in the physical, access control and network layer. We apply the framework on a representative 24-satellite constellation observing 5 global regions. Results show appropriately low latency in information exchange (average within 1/3$^{rd}$ available time for implicit consensus), enabling the onboard scheduler to observe ~7% more flood magnitude than a ground-based implementation. Both onboard and offline versions performed ~98% better than constellations without agility**.

## I. Introduction

Effective Earth Observation (EO) using spacecraft (or satellites) needs a wide spread of response and revisit frequencies, ranging from sub- hour for disaster monitoring to daily for meteorology to weekly for land cover monitoring. Satellites at altitudes below 900 km provide adequate spatial resolution but need large numbers for adequate response times. Small satellites (sub-180 kg) make constellations economically feasible to develop and launch. Adding (A) *full-body reorientation agility* to such small satellites, (B) *onboard prediction models* and (C) *onboard planning* to their software, and (D) *inter-satellite communication* to the constellation further improves science-driven responsiveness for any given number of satellites in any given orbits. Planning algorithms for science-driven, responsive EO have been traditionally developed for single, large satellite missions [1] and for constellations of large satellites [2], but they are ground-based and do not allow dynamic re-orientation. Constellation planning and scheduling for re-orientable imaging [3-5], using static or dynamic [6,7] assumptions on transition time, and downlink [8] have been done separately, or co-optimized for imaging and downlink [9-11]. However, such schedulers are ground-based and are not responsive to past observations using updated predictions. Recent Sensor Webs' planners that are responsive, to say riverine floods [29] and volcanos [30], are ground-based and cannot use inter-satellite links (ISLs) for coordination. Constellations that have shown onboard scheduled imaging and downlink aided by ISLs [12-14] are either not responsive using predictions or are not dynamically re-orientable. Optimizing all *four factors A-D* at constellation scale is, to our knowledge, sparse. Theoretical stochastic algorithms for multiple payloads [16] and satellite fleets [17] exist but are computationally intractable for flight. Onboard but greedy scheduling of agile satellite by leveraging ISLs for coordination has been attempted [18], however the planners use one-time predictions or trivial analytics (e.g. location seen?) instead of using observation-based predictions from science models with onboard-appropriate computational efficiency [19,20]. This paper introduces a *novel* planner that *functionally* improves the state-of-art by incorporating physics-based models in the predictor that are responsive to past observations while enabling dynamic spacecraft re-orientation and autonomous communication for coordination.

## II. Methodology

We describe an algorithmic framework that combines spacecraft physics models of orbital mechanics (OM), attitude control systems (ACS) and ISLs, along with an *autonomous, dynamic predictor and planner* to schedule the re-orientation of and observation by any satellite (sat) in a constellation of a set of ground regions with rapidly evolving phenomena. The proposed algorithm may run onboard satellites, which generate data bundles after executing scheduled observations and broadcast them using ISLs. ISLs are simulated using DARPA's delay/disruption tolerant network (DTN) protocol [21], a standard for routing in dynamic and intermittent operation environments. Bundles contain information about the ground points observed and meta-data from science models. Considering networking delays in a temporally varying disjoint graph, fast-changing access to regions (sat ground velocity ~7 km/s) and fast-changing phenomena, satellites are not expected to iterate on acknowledgments to

Sponsored by the NASA Earth Science Technology Office. *Affiliations listed are at the time of the project.

establish explicit consensus. Instead, we aim at implicit consensus; the more a satellite knows about a region before its observation opportunity, the better its scheduler performance. The algorithm may also run on the ground, i.e. the satellites can downlink their observed data, the ground will run the proposed algorithms, and uplink the resultant schedule to satellites. Since the ground stations are expected to be inter-connected on Earth and in sync with each other at all times, the optimization is centralized, and the resultant schedule avoids redundant observations due to lack of satellite consensus. However, since information relay occurs at only sat-ground contacts (dependent on orbits, ground network), the scheduler may use significantly *outdated* information compared to the distributed, onboard bundles. The transiency of the phenomenon being observed and robustness to latency in exchanging inferences determines if the onboard, decentralized implementation of our proposed algorithm is more effective than the ground, centralized one, or vice versa. The algorithmic framework and information flow is summarized in Figure 1 and detailed in Appendix A. The individual modules (OM [15, 34], ACS [22], communication [23]) are described in separate publications. This paper focuses on the scheduler based on Dynamic Programming (Appendix A.4) using module inputs (Appendix A.1) and the science prediction models (Appendix A.3) which assimilate satellite-observed precipitation into hydrologic models in real time to update flood forecasts [20].

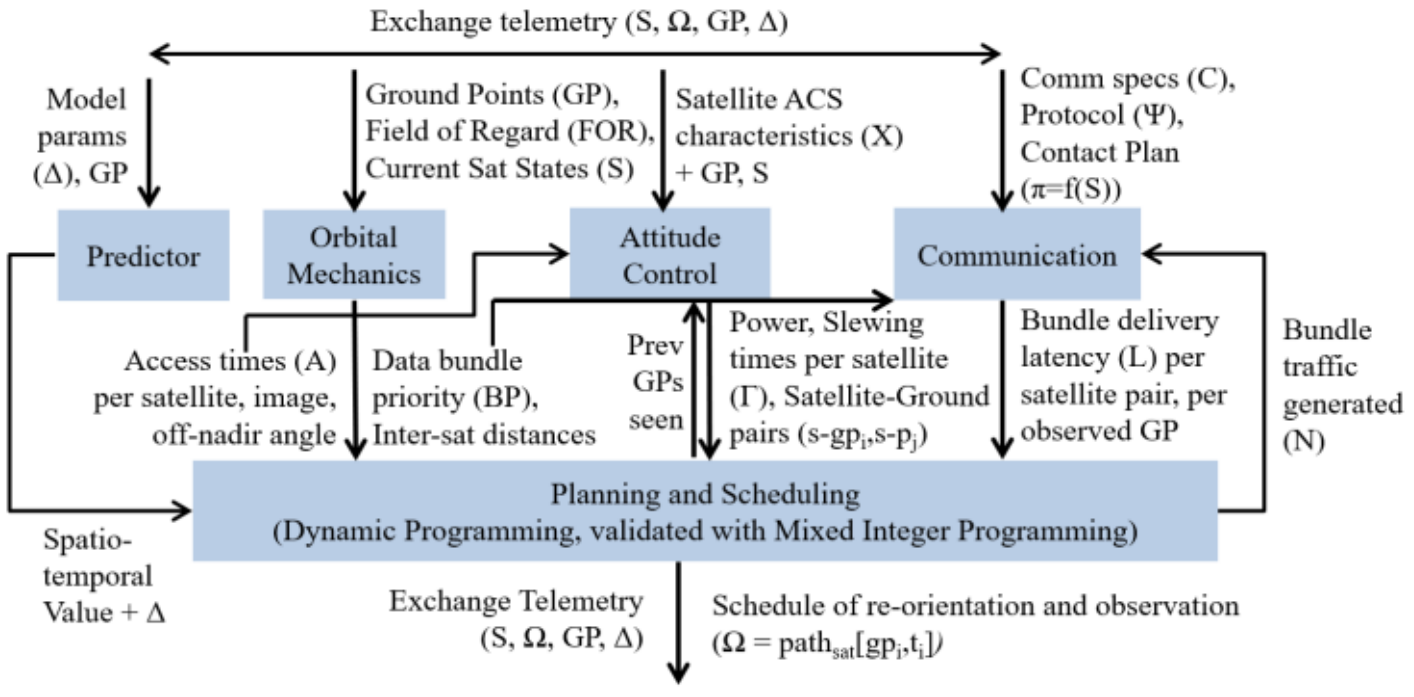

Figure 1 – Major information flows between the modules in the proposed agile EO framework, expected to run onboard every satellite in a constellation, applied to global urban flooding in this paper. This framework can exchange information (hence telemetry I/O) between the satellites as peer-to-peer or via the ground as reverse bent pipe architecture, to improve scheduling (output).

## III. Results from Full System Simulations

To evaluate the predictor and planner joint performance, we simulate a case study of 24 (20 kg cubic) satellites in a 3-plane Walker constellation (homogenous with 3 orbital planes of 8 satellites each) observing floods in 5 global regions over a 6-hour planning horizon. For the onboard planner, the latency of DTN delivery over all satellite pairs for all bundles was compared to the gaps experienced by any region between access by those same satellite pairs. For any pair of satellites with given bundle priority, if the longest latency is shorter than the shortest gap, each satellite can be considered fully updated with observations from all others. They can plan the same schedules for each other despite distributed compute on a disjoint graph. Simulations confirmed *successful comms*: All bundles were transmitted within gap time except a few lower priority satellite pairs [18, 23]. Median transmit time was less than 1/3$^{rd}$ median gap time. The DP-based scheduler maximizes cumulative value over all sat schedules (Figure 1), as the total observed flood magnitude by the constellation (Appendix B.3). The impact of varying planning horizon time on performance and runtimes for schedule execution (Appendix A.2) is shown in Figure 2. Runtime trades show *the scheduler is onboard-capable* and can be made efficient by executing smaller planning horizons without loss of value.

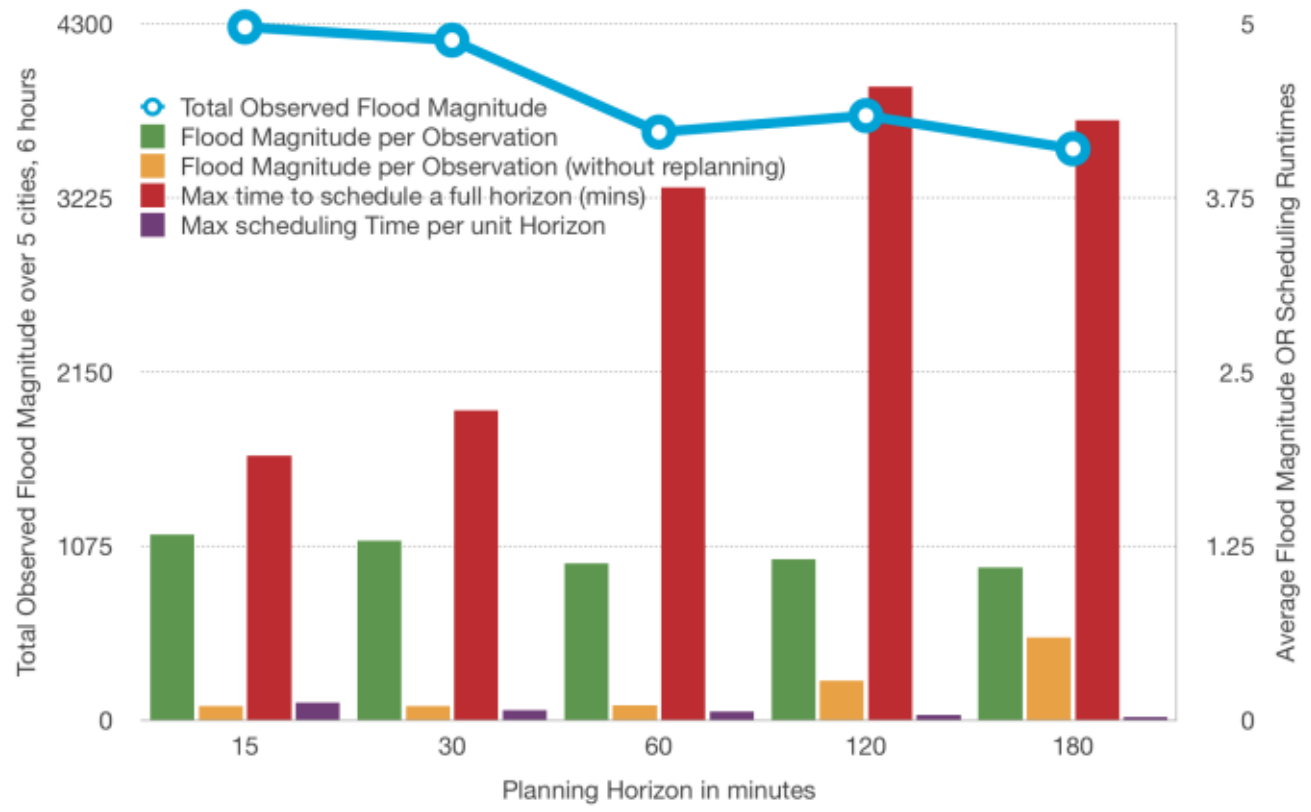

Figure 2—Observed flood magnitude - total (blue, left Y) and per observation (green, right Y) – both decrease with increasing planning horizon time and thus, decreasing re-planning frequency. Observed magnitude per observation (yellow, right Y) increases with horizon time for single plans. Scheduling runtime (red, right Y) increases with horizon time, but efficiency (purple, right Y) decreases.

We compared the performance of the framework when implemented onboard (at horizons of 5-15 mins, replanning every 3-10 mins) against on the ground (every 99-198 mins, corresponding to 7-4 GS contacts in the 6-hr sim) – results in Table 1. Best cases are bold-faced as recommended planner parameters, and each row compared to it as a % in brackets. Both scheduler implementations improved value over statically pointed radars on same satellites by 98% (total flood magnitude of 50.16 over 6h). Our performance eval methodology also provides a quantifiable way to decide if the benefit of agility (+98%), of onboard implementation (+7%), or recommended planner parameters (+5%) is worth the cost.

Table 1—Comparison of schedulers run Onboard vs. in Ground Stations (GS). Centralized planner's horizon is the time between GS contacts (99-198 mins). See Appendix B.3 for detailed discussion.

| | Planning Horizon (mins) | Total Observed Flood Magnitude | Flood Magnitude per Observation | Flood Magnitude per Observation (w/o replanning) | Max Time for each planHorizon (mins) | Max Time per unit planHorizon |
|---|---|---|---|---|---|---|
| Decen-tralized Plan Onboard before entering Region | 5 (3m replan) | 2661.7 (-1.6%) | 0.825 (-0.7%) | 0.1 (-83.1%) | **1.794 (best)** | 0.359 (+1336%) |
| | 10 (5m replan) | **2703.7 (best)** | **0.831 (best)** | 0.1 (-83.1%) | 2.374 (+32.3%) | 0.237 (+848%) |
| | 15 (10m replan) | 2638 (-2.43%) | 0.819 (-1.44%) | 0.1 (-83.1%) | 2.374 (+32.3) | 0.158 (+532%) |
| Central-ized Plan on Ground | GS contact every 50m | 2580.3 (-4.56%) | 0.799 (-3.85%) | 0.156 (-73.6%) | 3.705 (+107%) | 0.037 (+48%) |
| | GS contact every 99m | 2528.5 (-6.5%) | 0.787 (-5.3%) | **0.591 (best)** | 5.029 (+180%) | **0.025 (best)** |

## Acknowledgment

The authors thank Richard Levinson (NASA Ames Research Center) for his valuable review.

## Appendix

### A. Description of Algorithms in the Planner and Predictor

The proposed planner/scheduler ingests the time-varying outputs of the OM, ACS and Comm modules and uses them alongside dynamic science predictions to output the schedule $path_{sat}[gp_i,t_i]$, an array of tuples $[gp_i,t_i]$ which represents when ($t_i$) a *sat* should capture any $gp_i$. The executed schedule dictates the number of observations a satellite makes over any region, which dictates the number, size and timing of bundles generated for broadcast, therefore, we include a feedback loop between the scheduler and comm module. Slew characteristics depend on the previous GPs the satellite was observing as well as its next target, thus creating a feedback loop between the ACS and scheduler. Algorithmic details for all elements are described below. Note: The proposed framework is modular, and implementation of *any module* is generalizable.

#### 1) Ancillary Dynamic Inputs

The OM module [15, 34] propagates orbits in a constellation and computes possible coverage of known regions of interest (appropriately discretized into Grid Points (GPs)) within a Field of Regard (FOR is the angular area a sat *can* observe by re-orienting). It also provides line-of-sight (LOS) availability, inter-sat distances, and the priority of bundle delivery to the Comm/ISL module between every pair of satellites. For example, if Sat1 generates data over Dallas, and Sat2 is the next to access Dallas, Sat2 is given the highest priority. The comm module uses these priorities to schedule the DTN.

The ACS module computes the time required by any satellite to slew from one GP to another (including satellite orbital movement), resultant power, momentum and stabilization profiles. Sequential convex programming-based trajectory optimization is used to solve two types of problem formulations, as a function of the required objective: Minimum-Time Optimal Control Problem and Minimum Attitude-Error-and-Control-Effort Optimal Control Problem [22], and the output standardized for planner input.

The comms module [23] for onboard planning input computes the link budget for known specs (e.g. radio power, antenna size) in the physical layer. It uses the resultant data rate to simulate DTN in the network layer, and compute latency to deliver any bundle between any given pair of satellites. It assumes frequency division multiple access in the MAC layer. The comms module for ground-based planning is a simple uplink and downlink scheduler [8,11,33] modeled to ensure the required cadence of GS contacts (e.g. in Table 1).

#### 2) Scheduler Execution

The scheduler is run within a satellite's execution loop (Algorithm 1) that processes bundles received until time *tNow* through the DTN (line 6), and updates its knowledge of GPs observed by other sats *c* as broadcasted at $tSrc_c$ ($path_c[gp_i,t_i \leq tSrc_c]$) and corresponding measurements and estimates of the observed region ($modelParams(t_i \leq tSrc_c)$). It then computes schedules and makes observations (line 10-11), updates its internal science models such as those used for forecasts based on the new observations (line 12), broadcasts its updated model parameters as DTN bundles for every source-sink $sat \rightarrow c$ pair (line 14) across the constellation. Note that each DTN bundle is labeled with a pre-defined Time To Live (TTL) to prevent network flooding and increase chances of all bundles being delivered. A schedule can be computed for any time *tPlanNow* within a future *planHorizon* (after *tNow*) at any time *tNow* during execution (line 4), if it is before *planTimes_for_planHorizon* – a user-defined value which allows sufficient scheduler runtime (depends on compute hardware specs and *planHorizon*) and makespan [25]. The advantage of a DP-based schedule for *tPlanNow* is that it is additive to the previously planned schedules for *tNow* and observations are cumulative (line 9). See Figure 4 in Appendix A.4 for a visual representation of the Dynamic Programming outcome in line 8.

Algorithm 1—<u>Execution Algorithm</u>

1: Inputs – sat, $path_{sat}[gp_i,t_i]$
2: Output – $observations_{sat}[gp_i,t_i]$
3: **For** tNow in mission_simulation **do**
4:     **If** tNow in planTimes_for_planHorizon **then**//*Compute*
5:         **For** c in Constellation-{sat} **do**
6:             $modelParams(t_i \leq tSrc_c)$, $path_c[gp_i,t_i \leq tSrc_c] \leftarrow$ DTN_inbox(c,$tSrc_c$,sat,tNow)
7:         **Procedure** *Scheduling Algorithm* in Algorithm 4
8:             **return** $path_{sat}[gp_i,t_i \in planHorizon]$
9:         $path_{sat}[gp_i,t_i \geq tPlanNow] \leftarrow path_{sat}[gp_i,t_i \geq tPlanNow] + path_{sat}[gp_i,t_i \in planHorizon]$
10:     **Procedure** *execute* $path_{sat}[gp_i,t_i=tNow]$ //*Execute*
11:         **return** $observations_{sat}[gp_i,tNow]$
12:     $modelParams(t_i \leq tNow) \leftarrow observations_{sat}[gp_i, t_i \leq tNow]$
13:     **For** c in Constellation-{sat} **do** //*Communicate*
14:         DTN_outbox(sat,tNow, c,tNow+$TTL_c$) $\leftarrow$ $modelParams(t_i \leq tNow, path_{sat}[gp_i,t_i \leq tNow]$

#### 3) Quantifying Truth and Predictive Value

We apply the proposed framework to episodic precipitation and resultant urban floods, to demonstrate utility and scalability. We used the Dartmouth Flood Observatory [27] to study the frequency of global floods in 1985 – 2010, and identified 5 large cities (Dhaka, Sydney, Dallas, London, Rio de Janeiro) within flood-prone areas and defined watersheds around them in an 80km×80km area. The natural phenomena used as 'ground truth' was riverine flooding in the Atlanta metropolitan area for a storm event from April 5-6 2017, using the Weather Research and Forecasting (WRF) Hydrologic Model version 5 [28].

Our WRF-Hydro model used precipitation as the *dynamic input*, and was calibrated by adjusting *static parameters* (e.g. terrain, soils) until modeled streamflow matched measured flow at 9 USGS gauges, and validated on a June 20-21, 2017, rain event using the Nash-Sutcliffe Efficiency and Percent Bias metrics. These modeled flow rates were then normalized by the 2-year recurrence interval flow rate (Q2), i.e. the median annual flow estimated by USGS equations for an area that defines a 'flood' there, at 203 major tributary junctions (*trueFloodMag*$_{x=\{Tributary\ Junctions\},t}$ in Algorithm 2 line 4). The time($t_x$)-dependent, normalized flood magnitude Q/Q2 for every tributary junction is then assigned to all $gp_x$ within its contributing watershed, i.e. the area that drains to that junction. This represents *trueFlood*$_{init}$*([*$gp_x,t_y$*])*, and will be used to evaluate the overall performance per Algorithm 5. We simulated a 6-hour mission during which the 5 selected cities are 'flooded' at a resolution of 900m spatially and 15min temporally. The flood model local to Atlanta is intended as an example for all. Fidelity can be improved by using separate flood models per location to quantify initial expectations, predictive value and update functions.

### a) Initializing Predictive Value

Satellites will schedule observations based on expectations of precipitation, floods and thereby space-time observational *value*$_{init}$*([*$gp_x,t_y$*])* per Algorithm 2. For this paper, initial flood predictions (*floodMag*$_{init}$*([*$gp_x,t_y$*])*) were computed by a separate WRF-Hydro simulation (line 6) with the calibrated static parameters but a perturbed precipitation using Gaussian random fields [32, 20]. An average initial to true precipitation ratio of 0.52–1.2 resulted in initial to true flood ratio of 0.44–2.5, i.e. a significant difference in the expected vs. actual NWS flood stages (Figure 3). This initial predictive Q/Q2 was mapped using piecewise linear functions to 8-bit integer values, per Figure 3 and Equation (1).

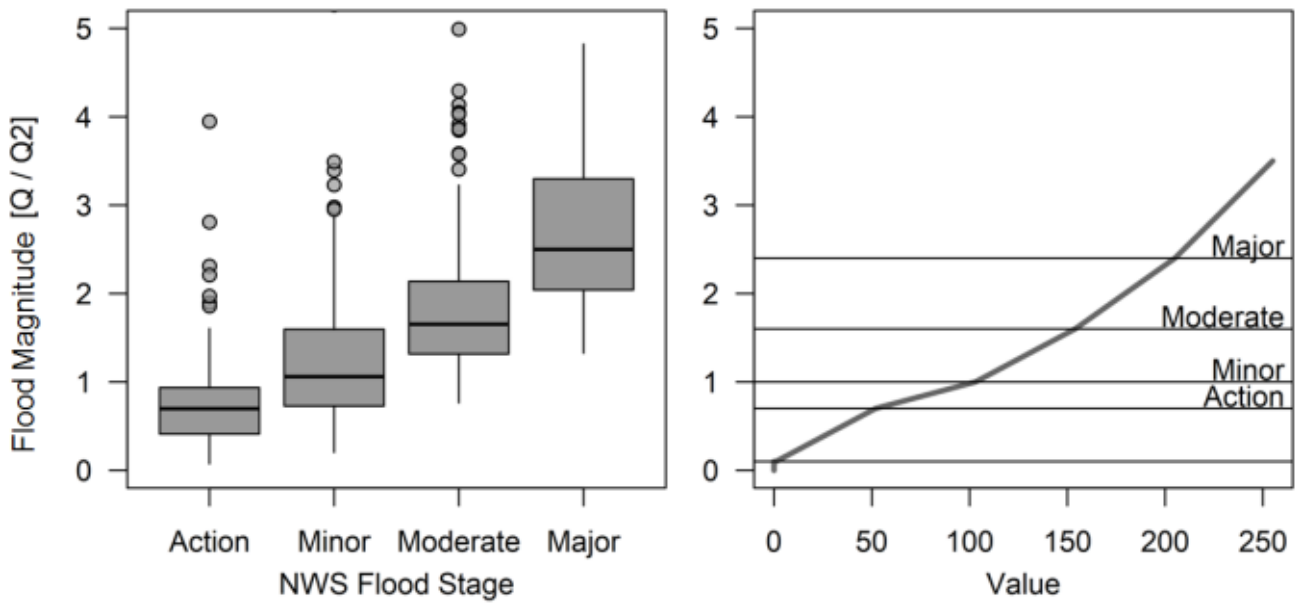

Figure 3—Boxplots showing values of Q/Q2 for USGS gages in Georgia with corresponding National Weather Service flood stages (left), which qualifies impacts to people and property. The median value from each flood stage (left) was used as a threshold to map Q/Q2 to an 8-bit scale called 'value' (right).

The value of observing a location increases with the severity of the disaster or the uncertainty around the geophysical variables associated with it. Areas with higher value correspond to watersheds with active flooding and inform the scheduler to prioritize satellite measurements of precipitation to determine if this flooding will worsen (with more rainfall) or abate. Future improvements to value functions are expected to include other environmental variables (e.g. cloud cover) and instrument dependencies (e.g. signal to noise ratio). Satellite instruments can *measure* space-time-varying precipitation, and onboard models can *estimate* (but not measure) floods. Thus, the scheduler starts with this predictive value map *value*$_{init}$*[*$gp_x,t_y$*]* and, as the constellation makes observations of precipitation over selected path$_c$[$gp_x,t_x$] by any satellite c, the value is updated to *value*$_{est}$*[*$gp_x,t_y$*]*. Both initial and updated (also called estimated) values are a function of initial and updated flood magnitudes respectively (Equation 1).

Algorithm 2—True Flood and Predictive Value Algorithm

1: Inputs – path$_c$[$gp_x,t_x$] $\forall c \in$ Constellation
2: Output – trueFlood([$gp_x,t_y$]), value$_{init}$[$gp_x,t_y$]
3: $Q_{x,t}$ = WRF-Hydro(staticParams,dynamicInputs$_{x,t}$)
4: trueFloodMag$_{x=\{Tributary\ Junctions\},t}$ = $Q_{x,t}$/Q2
5: trueFlood([$gp_x,t_y$]) ← trueFloodMag$_{x=\{Tributary\ Junctions\},t}$
6: floodMag$_{init}$([$gp_x,t_y$]) ← Line3-5 w/ diff(dynamicInputs$_{x,t}$)
7: **If** floodMag$_{init}$([$gp_x,t_y$])<0.1 **then**
8: floodMag$_{init}$([$gp_x,t_y$])=0.1
9: *compute* value$_{init}$[$gp_x,t_y$] per Equation (1)

(1)

{A,B} being piecewise linear coefficients derived from the following [value,floodMag] cuts in the curve by the National Weather Service (NWS) flood stages in Figure 3: [0.1,1.0], [0.7,52], [1,103], [1.6,154], [2.4,205].

### b) Updating Predictive Value

For every observation made at [$gp_{obs},t_{obs}$], the *value([*$gp_{obs}$*,* $t_y \geq t_{obs}$*])* is updated by an onboard predictive model, summarized in Algorithm 3.

Prediction outputs depend on *observations*$_{sat}$*([*$gp_{obs},t_{obs}$*])*, made by the satellite itself or derived from *modelParams(*$t_{obs}$*)* received from other satellites. For example, the latest precipitation measurements seen elsewhere can be inferred from model parameters received at the last broadcast *tSrc*$_c$ (line 4). While a higher fidelity, Long Short-Term Memory (LSTM) based prediction model was developed for this application [20], we used a simple regression-based version to demonstrate the framework in this paper. It was created by repeatedly running WRF-Hydro with varying precipitation, and a statistical analysis of the resulting stream discharge and flood magnitude performed. The change in peak flow was compared to the initial model, and a flood magnitude forecast derived using regression (*floodMag*$_{est}$ in Equation(2)). Line 9 thus estimates a predicted flood magnitude as a function of observed precipitation, updated owing to ISL comms, from the start of the 6h mission simulation. Kriging was used to estimate precipitation at all $\mathbf{gp_x}$ within the neighborhood δ of $\mathbf{gp_{obs}}$ and update all accordingly, albeit significant dependencies were found only within a 2 km radius. Per line 12, the updated value of any GP is also assumed to be a fraction (weight=1/number_of_times_seen) of its OSSE-provided value to encourage the scheduler to explore other GPs in the region and discover new flood targets, along with targeting known flooded GPs. Since the OSSE time resolution is 15min, once a GP is planned to be observed, subsequent observations within 15min are of zero value (equation 4).

Prediction outputs also depend on all the potential schedules (e.g. cartoon options in Figure 4) being evaluated by the sat running the predictor, called $potentPath_{sat}[gp_i,t_i]$, and the known schedules from all the other satellites in the constellation at the last broadcast, called $path_c[gp_i,t_i \leq tSrc_c]$. These schedules determine the expected precipitation that will be observed in the future (line 6-7).

For past and expected precipitation (or any geophysical variable) observations for past and expected schedules, the algorithm computes the updated value of those observations (line 12) and then re-computes the value $pathval[gp_i,t_i]$ for any potential (and actual) schedule $potentPath_{sat}[gp_i,t_i]$ ending at $[gp_i,t_i \geq tPlanNow]$ (line 13), given a known executed schedule by the other c satellites $path_c[gp_i,t_i \leq tSrc_c]$ in the constellation. The predictor thus scores all the schedule options to be then evaluated by the DP Scheduler (Algorithm 4). Note that Equation (3) and (5) are normalized over the GPs observed.

Algorithm 3—Value Update Algorithm

```
1: Inputs – sat, tPlanNow, floodMag_init([gp_x,t_y]),
   potentPath_sat[gp_i,t_i], path_c[gp_i,t_i≤tSrc_c], modelParams(t_i≤tSrc_c),
2: Output – pathval[sat][gp_x,t_y≥tPlanNow], precip_obs([gp_obs,t_obs])
3: For c in Constellation do
4:     truePrecip([gp_i,t_i]) ← modelParams(t_i≤tSrc_c) //latest data
5:     If c == sat then
6:         precip_obs([gp_obs,t_obs]) ←
                truePrecip([gp_i,t_i] ϵ potentPath_sat[gp_i,t_i])
7:     Else: precip_obs([gp_obs,t_obs]) ←
                truePrecip([gp_i,t_i] ϵ path_c[gp_i,t_i])
8:   Procedure Apply Equation (2)
9:       return floodMag_est([gp_x,t_y])  // updated estimate
10: For t ϵ [tPlanNow,planHorizon(end)] do
11:     compute value_est([gp_x,t_y]) per Equation (1)
12:     compute value_new([gp_x,t_y]) per Equation (3) and (4)
13:     compute pathval_sat([gp_i,t_i≥tPlanNow]) per Equation (5)
```

$$\begin{aligned}
&floodMag_{est}([gp_x \in \delta_{gp_{obs}}, t_y > t_{obs}]) \\
&= floodMag_{init}([gp_x \in \delta_{gp_{obs}}, t_y > t_{obs}]) + 1.85 \\
&* \left[\frac{\sum_{t=SimStart}^{t=t_{obs}} precip_{obs}([gp_{obs}, t])}{\sum_{t=simStart}^{t=t_{obs}} precip_{est}([gp_{obs}, t])} - 1\right] \\
&* floodMag_{init}([gp_x, t_y > t_{obs}]) \\
&\qquad \text{for } \frac{\sum_{t=simStart}^{t=t_{obs}} precip_{obs}([gp_{obs},t])}{\sum_{t=simStart}^{t=t_{obs}} precip_{est}([gp_{obs},t])} \geq 0.5 \\
&floodMag_{est}([gp_x \in \delta_{gp_{obs}}, t_y > t_{obs}]) \\
&= 0.05 * floodMag_{init}([gp_x \in \delta_{gp_{obs}}, t_y > t_{obs}]) \text{ , otherwise}
\end{aligned} \tag{2}$$

$$value_{new}([gp_x, t_y > t_{obs}]) = \frac{value_{est}([gp_x \in \delta_{gp_{obs}}, t_y > t_{obs}])}{\sum_{t=planHorizon(start)}^{t=tPlanNow} bool\{path_{sat}([gp_i == gp_x, t_i])\}} \tag{3}$$

$$\sum_{t=t_y}^{t=t_y+15m} value_{new}([gp_x, t]) = 0 \ \forall x \in [1,numGP] \tag{4}$$

$$pathval_{sat}([gp_i, t_i > t_{obs}]) = \frac{\sum_{[gp_x,t_y] in\ path_{sat}([gp_i,t_i])}^{pathlength\ of\ path_{sat}} value_{new}([gp_x \in \delta_{gp_{obs}}, t_y > t_{obs}])}{\sum_{t=planHorizon(start)}^{t=t_y} bool\{path_{sat}([gp_i, t_i])\}} \tag{5}$$

*4) Dynamic Programming based Scheduler*

The proposed scheduler (Algorithm 4) uses dynamic programming (DP) to greedily schedule a vector of tuples $[gp_i,t_i]\ \forall i \in [1,pathlength]$, for every sat. Each satellite is theorized to possess its own DP scheduler on board (or its own thread on ground) – cartoon version in Figure 4, where the nodes have varying $value_{init}([gp_x,t_y])$, $\forall x \in [1,numGP]$, $y \in [1,horiz_tSteps]$ within a region of interest and planning horizon. Any *path* has to trace a graph of time steps and GPs. For every time step *tPlanNow* (at 1s resolution within *planHorizon*), the scheduler steps through the GPs *gpNow* within the sat's FOR, and computes the cumulative value *pathval([gpNow,tPlanNow])* of each path ending at *[gpNow,tPlanNow]*, e.g. in Figure 4. Per DP, line 8 creates all potential paths ending at *[gpNow,tPlanNow]*, by adding the node to all paths ending at *[gpBef,tPlanBef]*, for all possible nodes *[gpBef,tPlanBef]* $\forall$ *gpBef*$\in$*[1,numGP], tPlanBef* $\in$*[ tPlanNow-max(slewTime), tPlanNow-min(slewTime)]*. The value of any potential path *pathval* is re-computed (line 9) using $floodMag_{init}$ at all $[gp_x,t_y]$ and the satellite's knowledge of the executed *observations* and derived *modelParams* by the rest of the constellation (Algorithm 1 line 6). If the scheduling sat's FOR at *tPlanNow* overlaps with any other's FOR, it must estimate value for all possible paths by the other *s*, and maintain cumulative value numbers for every permutation of paths among them in *satsWoverlappingFOR* (line 6), and choose the best permutation.

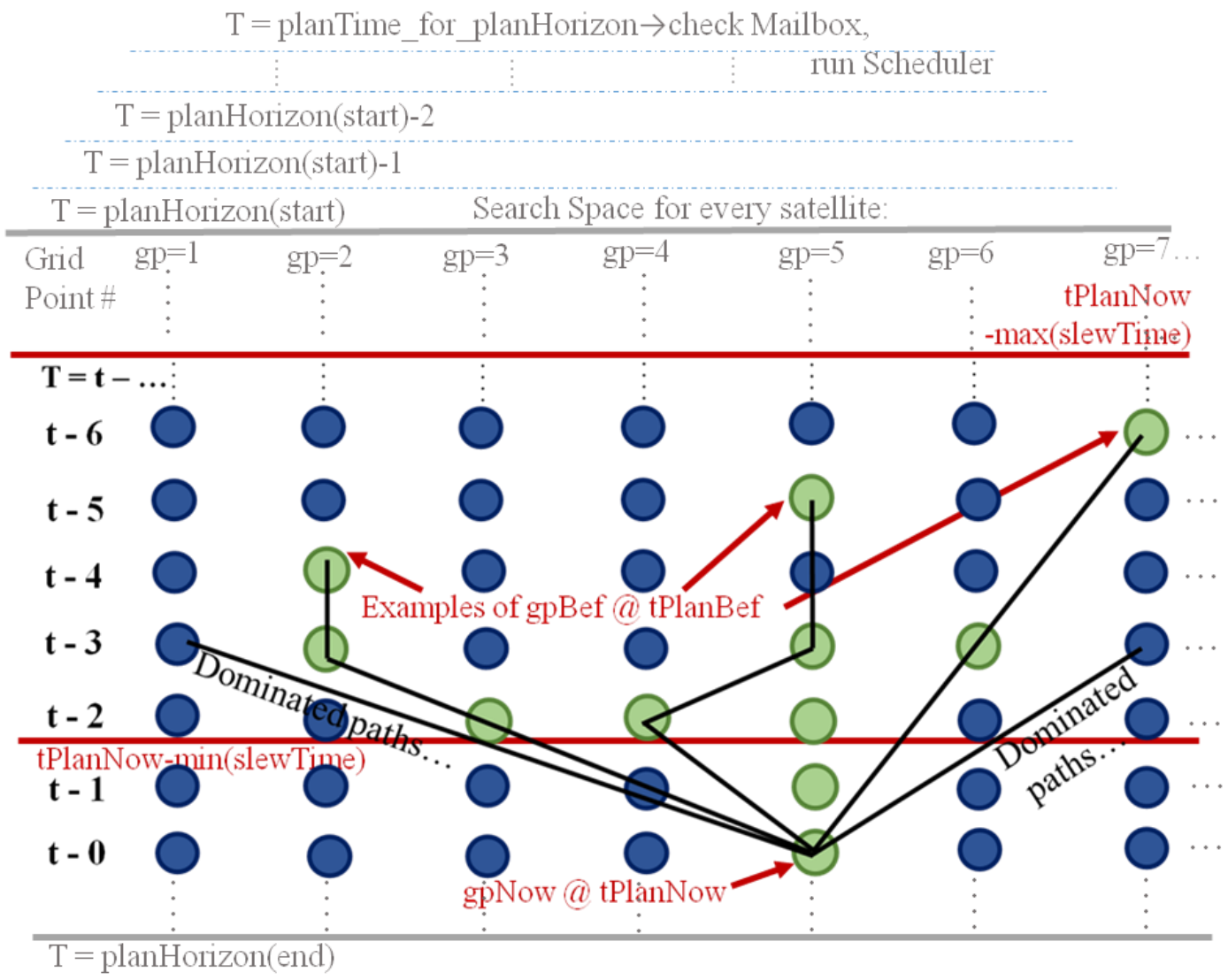

Figure 4—State space searched by any satellite's scheduler, within a pre-defined planning horizon (solid grey horizontal lines), to compute the optimum path ending at *[gpNow,tPlanNow]* assuming no FOR overlaps. Example non-dominated paths are seen as solid, black lines with green selected nodes. Only maneuvers beginning within solid red horizontal lines (called *Ground-PntsInBND*) are considered because they represent the min/max time required to end at *tPlanNow*. The scheduler steps through every *tPlanNow* in *planHorizon,* and considers every possible *gpNow.*

Starting with the paths with **maximum cumulative value** (line 11), *slewTime* of the last leg *[gpBef,tPlanBef]* → *[gpNow,tPlanNow]* (or combination of legs for overlapping sats) is computed. *slewTime* is the time required to re-orient the satellite from observing *gpBef* at *tPlanBef*, to *gpNow* at *tPlanNow,* and the output of the ADS module that models the spacecraft dynamics and produces a simple calculator for lookup [22]. Only node pairs whose slew times are within available access time constitute valid paths. The nodes between the red horizontal lines in Figure 4 are examples of *[gpBef,tPlanBef];* searching only a possibly valid portion of the space (e.g., *t-2* through *t-9*) mitigates computational load (line 5). For the first instance when the time required is shorter than allowed between the *tPlanBef→tPlanNow* gap, the path is stored as the optimum path $path_{sat}[gpNow,tPlanNow]$, and all other paths ending at *[gpNow,tPlanNow]* discarded (to prevent memory becoming astronomical). Future implementations may explore ways to preserve some dominated paths since globally optimum solutions (that are tied with locally optimum ones at *tPlanNow*) may be eliminated by pruning.

Algorithm 4—<u>Scheduling Algorithm</u>

```
1: Inputs – sat, planHorizon, floodMag_init([gp_x,t_y])
2: Output – path_sat[gp_i,t_i]
3: For tPlanNow in planHorizon do
4:    For gpNow in GroundPntsInFOR(sat, tPlanNow) do
5:       GroundPntsInBND=[tPlanNow-max(slewTime):
                        tPlanNow-min(slewTime),1:numGP]
6:       For s in satsWoverlappingFOR(sat,gpNow) do
7:          For [gpBef,tPlanBef] in GroundPntsInBND do
8:             potentPath_s[gpNow,tPlanNow]=
                  path_s[gpBef_s,tPlanBef_s]+[gpNow,tPlanNow]
9:             Procedure Value Update Algorithm in Algorithm 3
                  return pathval_s[gp_x,t_y≥tPlanNow]
10:         v_combi =
                  pathval[permute(s in satsWoverlappingFOR)]
11:         For vn in reverse_sort(v_combi)
12:            slewTime=computeManueverTimes(s_combi,
                     [gpBef_s_combi,tPlanBef_s_combi],
                     [gpNow,tPlanNow])
13:            If slewTime≤[tPlanNow-tPlanBef_s_combi]  then
14:               path_s[gpNow,tPlanNow]←
                  path_s[gpBef_s,tPlanBef_s]+[gpNow,tPlanNow]
15:               break // forLoop for vn
```

The advantages of this algorithm are runtime being linearly proportionate to the planning horizon; the scheduler can be stopped at any *tPlanNow* and yet have a valid plan; complex or hierarchical value functions with non-linear dependencies or multi-system interactions are easy to incorporate; and that complexity per sat per region is $O(n(T)\times n(GP)^{2\times n(S)})$, where n(S) is the number of satellites that can access the same GPs at the same time. If satellite FORs are non-overlapping, runtime or space complexity does not depend on the size of the constellation. Integer programming (IP) tests were able to verify that optimality of a similar algorithm for single satellites was within 10% [7]. The DP solution was 22% lower than the IP optimality bounds for constellations, which is well within the optimality bounds of greedy scheduling for unconstrained submodular functions [25]. However, DP schedules were found at nearly four orders of magnitude faster than IP, therefore far more suited for real time implementations.

*B. System Evaluation using Simulations*

The 24-satellite experimental set up summarized in Section 3 ensured that all satellites were simulated at a 710 km altitude, 98.5 deg inclination, circular orbits like Landsat. The 24-sat topology is the minimum nodes for continuous DTN to enable <6-hr urban flood monitoring. Each satellite carries a precipitation radar with an 8km footprint [24]. The FOR is chosen to limit the maximum off-nadir angle to 55 deg, because it corresponds to 5x distortion of the nadir ground resolution (science-driven limit to combine observations in any region). **Note:** This experiment is an example to demonstrate impact and trade-offs; the proposed framework is *generalizable* to any ISL-enabled, agile constellation where an appropriate prediction model is available to use data collected by *generalizable* instruments and then assimilated.

Our system eval closely aligns with NASA's science-driven evaluation of mission design or operational planning, which typically leverages Observing System Simulation Experiments (OSSEs) [31]. An OSSE comprises of a well-calibrated model or in-situ collected data used as the 'ground truth', from which 'synthetic observations' can be computed for any observing system as a function of instrument specs. Next, random errors representing measurement uncertainty are added to the 'observations' and used to forecast nature. The difference between synthetic observations and forecasts, i.e. a noisy subset of the ground truth, and the full ground truth assesses the effectiveness of instruments or operational plans.

$$updateValError = \sum_{x=1}^{numGP} \sum_{y=1}^{tSteps_in_model} \partial_{x,y}$$

where,

$$\partial_{x,y} = \left| \begin{array}{c} trueFlood([gp_x, t_y]) \\ -floodMag_{est}([gp_x, t_y]) \end{array} \right|$$

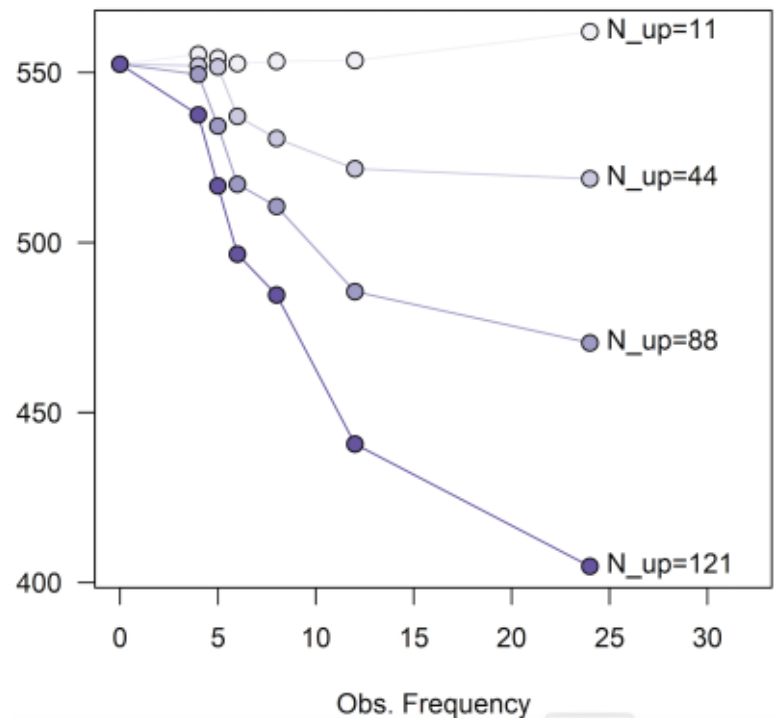

Figure 5—Validation the value update model where the Y-axis is the equation comparing estimated and true flood magnitude, which falls with increasing number of observations in space (#GP ~=50*N_up) and time (Obs. Frequency on X-axis = # of 15-min intervals sampled out of 24 in a 6 hr simulation period).

*1) Predictor Eval*

The prediction and value update model was validated by comparing the true flood (Algorithm 2 line 5) against estimated flood (Algorithm 3 line 9) for 28 hypothetical observation schedules. Results are seen in Figure 5. #N_up is the number of observations made, corresponding to ~50 GPs each. #Obs. Frequency is the number of times any region was sampled in 6 hours, assuming uniformly spread observations. The decrease in error is apparent as the observed GPs and observation time snapshots increases. Using a higher fidelity prediction model (e.g. LSTM [20]) is expected to improve results here and in the end-to-end system. Predictions can be further improved using updated machine learning methods.

### 2) *Communications Eval*

The communications module [23] used a small sat radio with 5W RF transmit power and yielded >1kbps effective data rate across OM-computed link distances. 8341 bundles of 2 kbits each were generated and sent over the DTN in the 6-hour simulation, and no bundles dropped within TTL. Simulations showed that latency improves with decreasing bundle size and traffic, and implicit consensus is possible at 500 bits. The Multi-Frequency Frequency Division Multiple Access (MF-FDMA) protocol was found to avoid MAC interference without any data-rate penalty. Details are available in Ref[23].

### 3) *Planner System Eval*

The planner is evaluated inclusive of all its inputs, i.e. predictor, comms, OM, and ADS, as structured in Figure 1. Algorithm 5 summarizes the evaluation process, which calls all algorithms previously listed.

Algorithm 5—Planner Performance Algorithm

1: **For** sat in Constellation *//parallel run on #sat threads* **do**
2: **Procedure** *Execution Algorithm* in Algorithm 1
3: **return** $\text{observations}_{sat}([gp_{obs}, t_{obs}])$
4: floodmagTotal = $\sum_{gp}\sum_{t} \text{trueFlood}([gp_x, t_y] == \text{path}_c[gp_{obs}, t_{obs}] \;\forall c \in \text{Constellation})$

Runtime for only the *scheduler* (i.e. to run Algorithm 4) per sat was ~1.7% of the planning horizon, evaluated on MATLAB installed in a Mac OSX v10.13 with a 2.6 GHz processor and 16 GB of 2400 MHz memory. C-code executions are typically 4x faster on power-limited small sat processors (e.g. ~100 Mz on Zedboard) compared to native MATLAB executions on the Mac. Optimization is improved by longer horizons *and* more frequent replans. However, frequent replanning overhead increases the ratio of scheduler execution time to horizon time (efficiency).

**Onboard vs. Ground discussion**: When the scheduler runs on onboard, re-planning frequency is constrained by onboard power or constraints on processing the bundles streaming in through DTN. Ground planning does not have such constraints but is limited by the cadence it can communicate schedules to the satellite for execution. If no re-planning is possible, centralized planning with longer horizons (~198 mins, i.e. once in two orbits) yield the maximum value (0.591), as seen in Table 1. Thus, it is *recommended* to extend the horizon for the ground planner to suit the comms cadence.

Since our precipitation and subsequent flood scenario is fast-changing, frequent replans with shorter horizons (as seen with shorter DTN latencies) improve total and per unit flood value because plans are based on more updated values. However, observational value plateaus out at 5-min replans as the cost of shorter planning horizons balance the diminishing returns of more frequent replans (Table 1). The O(n) scheduler runtime also increases with planning horizon—1.8min for the best DTN case to 5min for the worst GS case. Thus, 10 min horizons re-planned every 5 mins is *recommended* for the onboard planner.

Run-time fraction of the planning horizon improves from 36% of a 15-min horizon to 2.5% of a 180-min horizon because frequent replans and shorter horizons add significant overhead of mailbox checking, bundle processing and value re-computation (Figure 2, purple bars). Our performance eval methodology also provides a quantifiable way to decide if the benefit of additional value for the onboard planner parameters recommended above is worth the additional cost of run-time compute. For example, relative value can be assessed as ~5% more flood magnitude (Table 1) and disproportionately higher increase in GPs observed for each NWS flood category (action/minor/moderate) with replan frequency.

The ground-based implementation of the scheduler has the advantage that all the GS can be immediately sync-ed. There is no risk of overlapping observations (if DTN bundles don't reach in time) and all satellites know each other's centralized schedules. However, the disadvantage is ground-to-space latency. Even when each satellite gets 2 GS contacts per orbit (i.e. 30/day), value updates are from at least an orbit earlier due to collection-uplink-reschedule-downlink latency between any satellite pair and the total and unit flood-magnitude observed is ~3% less than any DTN-enabled decentralized run. Satellites with 1 GS contact per 99min orbit (15/day) are ~7% worse. Typically, small satellite missions commit to 2 contacts per day at NASA and 4-5 per day commercially, thus the benchmarked scenario expects utilization of dense GS infrastructure like Amazon Web Services. Thus, onboard implementations are *recommended* over ground for phenomena that evolve faster (like wildfire spread over snow melting), satellites whose orbits are at higher inclinations from the Equator, and regions of interest that are more equatorial (due to GS-space link opportunities becoming sparser). Our performance eval methodology also provides a quantifiable way to decide if the benefit of agility (+98%) or of onboard implementation (+7%) is worth the cost.

**Future Work:** While this paper demonstrates value of the proposed framework by incorporating a simple algorithmic predictive model, using higher fidelity nature runs and predictive models (see [20]) instead are expected to further improve planner outcomes as well as observed value.

We will also improve the runtime efficiency of the scheduler while continuing to optimize conflicting variables and performance plateaus. For example, see Reference [33] for a different implementation of a ground-based planner for agile satellites with responsive science feedback. The improved computational efficiency is amenable for onboard implementation, pending the incorporation of DTN bundle scheduling, such that latencies can be improved by managing congestion better. We will explore scheduler runs combining ground and onboard approaches because the right balance is a function of satellite orbits, infrastructure available, phenomena transiency and geographic locations of regions.